\documentclass[10pt,twocolumn,letterpaper]{article}

\usepackage{cvpr}
\usepackage{times}
\usepackage{epsfig}
\usepackage{graphicx}
\usepackage{amsmath}
\usepackage{amssymb}

\usepackage[pagebackref=true,breaklinks=true,letterpaper=true,colorlinks,bookmarks=false]{hyperref}

\cvprfinalcopy 

\def\cvprPaperID{1854} 

\ifcvprfinal\fi
\begin{document}

\title{A Comparative Study of Algorithms for Realtime Panoramic Video Blending}

\author{Zhe Zhu\\
TNList, Tsinghua University\\
Beijing, China\\
{\tt\small ajex1988@gmail.com}
\and
Jiaming Lu\\
TNList, Tsinghua University\\
Beijing, China\\
{\tt\small loyavefoever@gmail.com}
\and
Minxuan Wang\\
TNList, Tsinghua University\\
Beijing, China\\
{\tt\small hitminxuanwang@gmail.com}
\and
Songhai Zhang\\
TNList, Tsinghua University\\
Beijing, China\\
{\tt\small shz@tsinghua.edu.cn}
\and
Ralph R Martin\\
Cardiff University\\
Cardiff, UK\\
{\tt\small Ralph.Martin@cs.cardiff.ac.uk}
\and
Hantao Liu\\
Cardiff University\\
Cardiff, UK\\
{\tt\small LiuH35@cardiff.ac.uk}
\and
Shimin Hu\\
TNList, Tsinghua University\\
Beijing, China\\
{\tt\small shimin@tsinghua.edu.cn}
}
\maketitle

\begin{abstract}
   Unlike image blending algorithms, video blending algorithms have been little studied. In this paper, we investigate 6 popular blending algorithms---feather blending, multi-band blending, modified Poisson blending, mean value coordinate blending, multi-spline blending and convolution pyramid blending. We consider in particular realtime panoramic video blending, a key problem in various virtual reality tasks. To evaluate the performance of the 6 algorithms on this problem, we have created a video benchmark of several videos captured under various conditions. We analyze the time and memory needed by the above 6  algorithms, for both CPU and GPU implementations (where readily parallelizable). The visual quality provided by these algorithms is also evaluated both objectively and subjectively. The video benchmark and algorithm implementations are publicly available.\footnote {http://cg.cs.tsinghua.edu.cn/blending/}
\end{abstract}

\section{Introduction}
Many image editing~\cite{7272134} tasks involve blending, e.g., panorama stitching, and copy-and-pasting of objects into images. As human eyes are sensitive to color and lighting inconsistencies within images, image blending is used to provide  smooth transitions between image parts from different sources. Indeed, image blending is a standard part of modern image editing tools such as Adobe Photoshop.

While state-of-the-art image blending algorithms~\cite{agarwala2007efficient,burt1983a,Farbman:2011:CP:2070781.2024209,farbman2009coordinates,perez2003poisson,Tanaka:2012:SIC:2407156.2407173}
can achieve good results, it is hard to find evaluations of the trade-off between their efficiency and quality of results. This is mainly because these algorithms can provide high quality results in a short time: e.g.\  mean value coordinate blending~\cite{farbman2009coordinates} takes about 1~s in total to blend a region with 1 million pixels.

For video blending, especially at high resolution, the situation changes. The quantity of data is much larger than in images, so efficiency becomes a major concern. For example, virtual reality applications, e.g.\ involving live sports, can demand real-time content creation based on $360^\circ$ panoramic video blending; these panoramic videos are much larger than ordinary videos. In a typical 4k $360^\circ$ 30fps panoramic video, blending must be done in under 30~ms (and indeed rather less to allow time for other processing tasks on each frame). Thus, real-time high resolution video blending is much more challenging than image blending, and indeed, parallelization is often needed.

The aim of this paper is to compare various the suitability of various image blending algorithms for real-time usage for video blending in high resolution panoramic video stitching. We first briefly describe each algorithm, and analyse the relationships between them. Then, we conduct experiments on a benchmark data set, evaluating both their performance on different kinds of scenes, considering both time and memory costs, and the quality of the blended results, using both objective and subjective assessments. Unlike image blending, which is a one-shot operation, video blending considers successive frames, which may e.g.\  share common fixed camera positions. Some algorithms take advantage of this by a possibly lengthy precomputation; for short video clips, for methods like mean value coordinate blending, this may even take the majority of the computation time.

We do not include content-aware blending algorithms~\cite{jia2006drag-and-drop,tao2013error-tolerant} in our comparison as they are unsuited to real-time video blending, for two reasons: these methods are relatively slow due to the need to analyze content, and furthermore cannot readily ensure interframe coherence.

We have captured a set of benchmark videos including various types of scenes, to enable evaluation of different aspects of the blending algorithms. Each video has 6 separate streams; we also provide a stitching template which defines the positional relationships between the pixels in each stream, and those in the final panorama. A blending algorithm under test uses this to produce the panoramic result.

The contributions of this paper are:
\begin{itemize}
  \item Our publicly available benchmark set of videos for the evaluation of panoramic video blending.
  \item A comparative study of the suitability of several state-of-the-art image blending algorithms for panoramic video blending, which makes clear the advantages and disadvantages of each algorithm, as well as the relationships between them. Implementations of these algorithms are also publicly available.
\end{itemize}

In Section~\ref{sec:bpes} we describe the benchmark. We described the different blending algorithms and their relationships in Section~\ref{sec:blend}. The behaviour of these algorithms on our benchmark is examined in Section~\ref{sec:exp}, and we give our conclusions in Section~\ref{sec:conclusion}.

\section{Benchmark and Experimental Setting}
\label{sec:bpes}
\subsection{Formulation}
\label{seq:formulation}
\begin{figure}[t!]
\begin{tabular}{c c c}
  \includegraphics[width=0.3\linewidth]{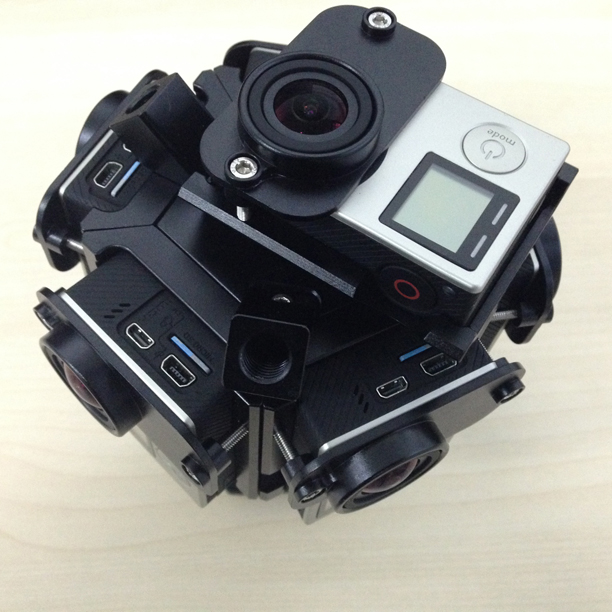} & \includegraphics[width=0.3\linewidth]{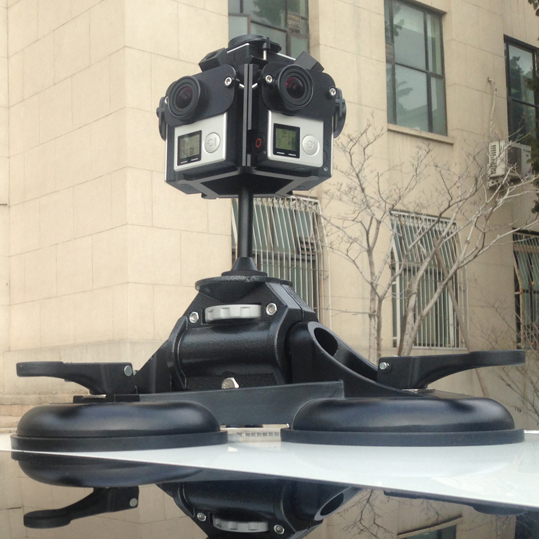} & \includegraphics[width=0.3\linewidth]{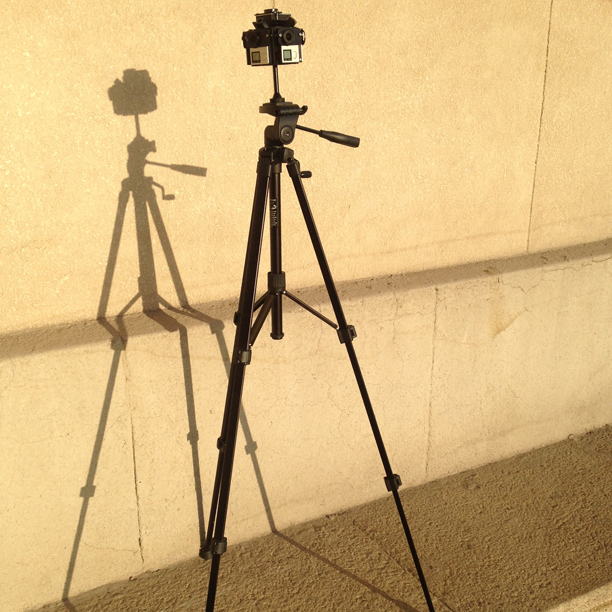}\\
\end{tabular}
\centering
   \caption{Capture device. Left: camera rig. Center: rig mounted on a car. Right: rig mounted on a tripod.}
\label{fig:rig}
\end{figure}

\begin{figure}[t!]
\includegraphics[width=1.0\linewidth]{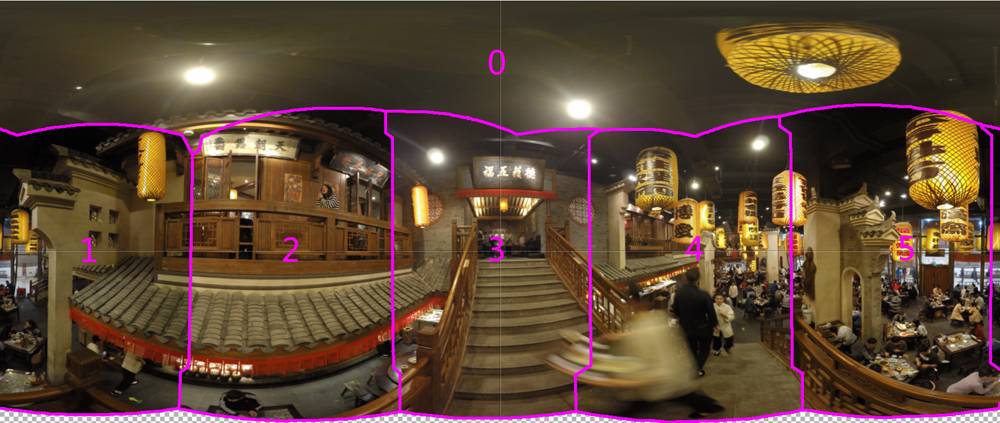}
\caption{A typical stitched panorama. Region 0 is captured by the upwards-pointing camera. Regions 1--5 are captured by the other cameras. Red lines indicate the fixed boundary seams between neighbouring video streams.}
\label{fig:template}
\end{figure}

As a basis for the panoramic video, $n$ video streams ${S_i}$, $i =1,\dots,n$, are recorded simultaneously with the same resolution ($n=6$ in our setup). Before blending, these must be stitched into a single frame of reference to form a panorama. Given our fixed camera rig, and known camera intrinsic parameters, we first perform radial distortion correction for each stream and match keypoints
 between neighbourhood streams. We pick one frame as a reference for each stream and set the correction to all the following frames. In this way we can ensure the coherence between frames. By choosing one stream as a base, we rotate other streams in the viewing sphere according to the best fitted yaw, roll and pitch angle  obtained from the matched keypoints.
Spherical projection is then used to map the rotated content in the viewing sphere to the planar panoramic output video. We then perform a local varying warp following~\cite{Zaragoza:2014:AIS:2693344.2693388} to further align the details. For example the stream of the top camera in Figure~\ref{fig:rig} is mapped to the top region in Figure~\ref{fig:template}.
The resulting pixels in the panorama corresponding to each initial stream are determined by a mapping function:
 \begin{equation}\label{equ:mapping}
{P_i} = \varphi_i ({S_i})
\end{equation}
We associate a mask $M_i$ for each mapped stream $P_i$, which contains 1 for pixels covered, and 0 for pixels not covered, in the output. Each mapped stream overlaps neighboring streams by about 20\% of its total area, providing the necessary data for those blending algorithms that require overlapping regions.
Other blending algorithms require boundaries between streams, which we determine in the overlap region  using distance transformations~\cite{v008a023} on the first frame
to find the locations equidistant to the corresponding streams. This boundary is used to trim the original mask $M_i$ to a new mask  $M_i'$.

We may put blending algorithms into two categories: those that calculate the blended pixels directly,  and those that first compute an offset map, and then add the offset map to the original video. The offset map is an image with the same resolution as each video frame, in which the pixel value at each position is the difference between the desired blended value and the original value.
Algorithms in the first category  obtain the final panoramic video $P$ by computing:
\begin{equation}\label{equ:formulation1}
P = f({P_1},{M_1},\dots,{P_n},{M_n})
\end{equation}
where $f$ is some function that performs operations on the mapped streams ${P_1},\dots,{P_n}$.
Algorithms in the second category produce the final panoramic video by computing:
\begin{equation}\label{equ:formulation2}
P = P' + P^*
\end{equation}
where $P'$ is a video obtained by directly trimming and compositing the mapped candidate streams along precomputed boundaries,
and $P^*$ is a combined offset map formed from the offset maps of each mapped candidate stream, using the same boundaries.
Thus $P'$ is computed using:
\begin{equation}\label{equ:p1}
P' = \sum\limits_{i = 1}^n {{M_i'}{P_i}'}
\end{equation}
where $P_i'$ is the mapped $i^{th}$ stream.
$P^*$ is defined in a similar way:
\begin{equation}\label{equ:p2}
P^* = \sum\limits_{i = 1}^n {{M_i'}{P_i}^*}.
\end{equation}
The way in which the individual offset maps $P^*_i$ are computed varies according to the blending algorithm.

\subsection{Benchmark}
\label{sec:benchmark}
To evaluate the performance of blending algorithms when used for panoramic video blending, we captured video data from various indoor and outdoor scenes, using a camera mount based on a six GoPro camera rig---see Figure~\ref{fig:rig}. Five cameras are arranged symmetrically in a plane around a vertical axis, while the last camera points vertically upwards. They are mapped to the panorama as shown in Figure~\ref{fig:template}. A GoPro Smart Remote is used to synchronize video capture from all cameras.

We captured videos with variation in three key properties---illuminance conditions, camera motion, and object distance to the camera---as changing them can significantly effect blending results.
Illuminance variations cover both  indoor and outdoor scenes with adequate and poor lighting. While video cameras often automatically determine exposure, changes in illuminance conditions may have a strong effect on the brightness of the videos.
 We provide videos from both static and moving camera setups. The latter causes content change along the boundary seams, which may be substantial e.g.\  if the camera rig is mounted on a moving vehicle. For video blending, we pick one frame as reference and compute a stitching template, and apply this template to all the other frames. This can ensure the coherence in the blended video. On the other hand, even we perfectly stitch the reference frame, when applied to other frames, the computed template can still cause misalignments. Thus
distances of the key objects to the camera in the video can also affect the blending results; objects with varying distances can cause bleeding artefacts near seams. As human eyes find larger objects more salient,  artefacts in objects closer to the cameras are often more obvious.
We have captured scenes with moving objects at near, intermediate and far distances.
In total, we have 4 illumination conditions, 2 motion types, and 3 distance types, giving 24 types of video; we provide 2 of each type giving  48 different scenes altogether, each lasting from several minutes to tens of minutes.

Each captured video stream has resolution $1920 \times 1440$ at a frame rate of 30~fps. The panoramic video has resolution $4000 \times 2000$.

\section{Blending Algorithms}
\label{sec:blend}
\subsection{Overview}
Image blending is well studied. Perhaps the most widely used approach is multi-band blending~\cite{burt1983a}. It is easy to implement and provides stable blending results. It  blends the images at each level of a Laplacian pyramid, and merges them to give the result. Perez et al~\cite{perez2003poisson} formulate image blending via a Poisson equation whose solution can be obtained by solving a large sparse linear system. Although this is mathematically elegant and provides  perfect results when the boundary is smooth, it is time consuming, especially for large images. It also suffers from bleeding artefacts when the blending boundary is insufficiently smooth. Agarwala~\cite{agarwala2007efficient} observes that the offset between the original content and the blended content in the target region is piecewise smooth, allowing ready approximation of the whole offset field by a quadtree. This significantly reduces the number of variables in the linear system, accelerating blending. Szeliski et al~\cite{export:144582} further observe that if each image has a separate offset field  represented by a low-dimensional spline, each offset field is everywhere smooth, not just piecewise smooth. As the spline has low dimensionality, the number of variables is further reduced. To avoid solving linear equations, Farbman et al~\cite{farbman2009coordinates} instead use mean-value coordinates (MVC) to interpolate the smooth offset field from boundary differences. For a target region of fixed shape, these coordinates can be precomputed and re-used for all frames. Furthermore, this method is readily parallelizable, but since it approximates the Poisson formulation, it too suffers from bleeding artefacts. In~\cite{Farbman:2011:CP:2070781.2024209} Farbman et al, observe that the key operation in MVC interpolation are convolution operations with large kernels; these can be approximated by several small kernels to further reduce computation. Poisson blending can also be improved by adding an intensity constraint~\cite{Tanaka:2012:SIC:2407156.2407173}, as explained later (and henceforth referred to as the \emph{modified Poisson} approach).

We analyze six representative blending algorithms, chosen for the following reasons. Feather blending has the lowest computational expense (apart from trivially mutually clipping the images), and provides a basic degree of visual quality.
 Multi-band blending is the most widely used approach in the open source community, and is relatively insensitive to misalignment.
  MVC blending can be readily parallelized, and avoids large linear equationa, while providing almost visually identical result to standard Poisson blending. Using a convolution pyramid approximates the MVC approach and further speeds it up.  Multi-spline blending uses another strategy to approximate the original Poisson equation, resulting in a significantly smaller linear system. Differences in formulation of modified Poisson blending lead to visually different blending results.

  We do not consider the original Poisson blending method, which is both slow and memory hungry, so unsuited to high resolution realtime video blending. We also do not consider the quadtree approximation to Poisson blending as it uses the smoothness of the offset map in a similar way to multi-spline blending, but the latter solves a smaller linear system.

  We now describe these algorithms in more detail.

\textbf{Feather Blending (FB)}: This simply linearly combines the two regions to be blended, using:
\begin{equation}\label{equ:fb}
P = \sum\limits_{i = 1}^n {{\omega _i}{P_i}}
\end{equation}
where $\omega_i$ is a per pixel weight map for each input stream. At each pixel, the weights of all streams sum to 1, so feathering only affects the overlap region. The simplest approach uses weights of 0.5 everywhere in the blend. A better approach equally weights the streams at their common boundary, with weights falling off the further we go into the opposite region, until they become zero.
As pixel value is independent, feathering is fully parallelizable.

\textbf{Multi-band Blending (MBB)}: This basically performs feather blending on images of different frequencies. Aa Laplacian pyramid is built, and the regions to be blended are linearly combined at each level. The final result is obtained by adding all blended images from the different levels. The Laplacian pyramid can be constructed in parallel using  equivalent
weighting functions~\cite{burt1983a}. As each level of the pyramid can be regarded as a function of the original image, it is possible to precompute the function mapping between the input image and the other levels, allowing computation of each level of the pyramid simultaneously. Combination of the Laplacian images using a Gaussian weight image is also fully parallelizable. Multi-band blending can be defined as:
\begin{equation}\label{equ:mbb1}
P = \sum\limits_{j = 1}^l {\textrm{EXPAND}({Q_j})},
\end{equation}
where $l$ is the number of layers of the pyramid, and $\textrm{EXPAND}()$ up-samples an image to the original resolution. $Q_j$ is defined as:
\begin{equation}\label{equ:mbb2}
{Q_j} = \sum\limits_{i = 1}^n {G_i^j} L_i^j,
\end{equation}
where $G_i^j$ is the $i^{th}$ stream's Gaussian pyramid at level $j$, and $L_i^j$ is the $i^{th}$ stream's Laplacian pyramid at level $l$.

\textbf{MVC Blending (MVCB)}: This approximates the Laplacian membrane used in Poisson blending, constructing a harmonic interpolant from the boundary intensity differences. Unlike Poisson blending which finds the final pixel values directly, MVC blending computes an offset map; the final blended result is obtained by adding this offset map to the region to be blended (see Equation~\ref{equ:formulation2}). Given a point $x$ in the region to be blended, $P^*(x)$ can be calculated by:
\begin{equation}\label{equ:mvc1}
{P^*}(x) = \sum\limits_{i = 0}^{m - 1} {{\lambda _i}(x)\mathrm{sub}({p_i})},
\end{equation}
where $p_i$ is some pixel along the boundary of the region to be blended, and $sub()$ is the difference operation between the two image regions at the same position(suppose we want to blend $T_a$ and $T_b$, and $T_b$ changes to fit $T_a$, $sub()$ performs $T_a-T_b$), and $m$ is the number of boundary points.
 $\lambda_i$ is the mean value coordinate of $x$ with respect to the boundary points---see~\cite{farbman2009coordinates}.
 For each pixel in the output region, the offset value is a weighted linear combination of the boundary differences using a combination weight derived from the pixel's mean value coordinate. As the boundary seams have fixed locations, the mean value coordinates and weights can be pre-computed once for all frames, saving effort for video blending. Since the value at each position of the offset map only depends on the boundary differences, MVCB is parallelizable.

\textbf{Convolution Pyramid Blending(CPB)}:
In MVCB, the final membrane(offset map)
 can be written as:
\begin{equation}\label{equ:cpb}
{P^*}(x) = \frac{{\sum\limits_k {{w_k}(x)b({x_k})} }}{{\sum\limits_k {{w_k}(x)} }},
\end{equation}
where $x_k$ are boundary points, $b(x)$ are boundary values and $w_k(x)$ are corresponding MVC coordinates. Following~\cite{Farbman:2011:CP:2070781.2024209}, Equation~\ref{equ:cpb} can be rewritten as a ratio of convolutions by incorporating a characteristic function ${\chi_{\hat P}}$ which is 1 where $\hat P$ is non-zero and 0 otherwise:
\begin{equation}\label{equ:cpb2}
{P^*}(x_i) = \frac{{\sum\limits_{j = 0}^n {w({x_i},{x_j})\hat P({x_j})} }}{{\sum\limits_{k = 0}^n {w({x_i},{x_j}){\chi _{\hat P}}({x_j})} }} = \frac{{w*\hat P}}{{w*{\chi _{\hat P}}}},
\end{equation}
where $\hat P$ is an extension of the boundary $b$ to the entire domain:
\begin{equation}\label{equ:cpb3}
\hat P({x_i}) =
    \begin{cases}
      b({x_k}), & \text{if}\ {x_i} = {x_k} \\
      0, & \text{otherwise}
    \end{cases}.
\end{equation}
Calculation of the offset map now involves convolutions with large filters. \emph{Multiscale transforms}~\cite{Farbman:2011:CP:2070781.2024209} allow these to be approximated by a set of small filters in linear time.

\begin{table*}
\centering
\caption{Computation times(per frame)
and memory usage for 4000$\times$2000 resolution, for the six algorithms.}
\label{tab:t1}
\begin{tabular}{|c|c|c|c|c|c|c|c|c|c|c|}
  \hline
    & FB  & FB(GPU)& MBB &MBB (GPU)& MVCB & MVCB (GPU)& CPB &CPB (GPU)& MSB & MPB \\
    \hline
  Mem (MB) & 498 & 428&  841 &2274 &3303 & 3982 &  942 & 2380  & 2225 & 1295  \\
  \hline
  Time (ms) & 1245  & 7& 4992  & 25 & 2535& 31& 4782 & 63  &  7940 &  11322   \\
  \hline
\end{tabular}
\end{table*}

\textbf{Multi-Spline Blending (MSB)}:
Using an energy minimization formulation~\cite{perez2003poisson}, Poisson blending  can be written in offset map form as:
\begin{equation}\label{equ:msb2}
\begin{split}
{E} = {\sum\limits_{i,j} {({P^*}_{i + 1,j}^{{l_{i + 1,j}}} - {P^*}_{i,j}^{{l_{i,j}}} - \hat g_{i,j}^x)} ^2} + \\ {({P^*}_{i,j + 1}^{{l_{i,j + 1}}} - {P^*}_{i,j}^{{l_{i,j}}} - \hat g_{i,j}^y)^2},
\end{split}
\end{equation}
where  $l_{i,j}$ indicates which stream each pixel comes from($(i,j)$ indicates the location in image plane, and the label can be obtained by the mask),
and the  (modified) gradient $\hat g_{i,j}^x$ is defined as:
\begin{equation}\label{equ:msb3}
\hat g_{i,j}^x = {P'}_{i,j}^{{l_{i,j}}} - {P'}_{i,j}^{{l_{i + 1,j}}} + {P'}_{i + 1,j}^{{l_{i,j}}} - {P'}_{i + 1,j}^{{l_{i + 1,j}}}
\end{equation}
where ${P'}_{i,j}^{l_{i,j}}$ is the pixel intensity at location $(i,j)$ choosing $l_{i,j}$th stream;
the modified $y$ gradient $\hat g_{i,j}^y$ is defined  similarly.
The energy $E$ can be minimized by solving a linear equation $Az=b$
 where $z$ represents the unknown pixel values in the offset map. By using spline cells
 to approximate the assumed smooth offset map, each pixel in the final offset map can be represented by:
\begin{equation}\label{equ:msb4}
{P^*}_{i,j}^l = \sum\limits_{k,m} {c_{k,m}^lB(i - kR,j - mR)},
\end{equation}
where $R$
is the pixel spacing (we choose 64 in our experiment) of the spline cells, $B(i - kR, j - mR)$ give the spline basis and $c_{k,m}$ are spline control points. In this way, the dimension of the linear system is reduced significantly.

\textbf{Modified Poisson Blending (MPB)} Tanaka et al~\cite{Tanaka:2012:SIC:2407156.2407173} modified the original Poisson energy function by adding an intensity constraint:
\begin{equation}\label{equ:mpb1}
E' = \sum\limits_{i,j} {\varepsilon {{({I_{i,j}} - {P_{i,j}})}^2} + {{({g_{i,j}} - \nabla {P_{i,j}})}^2}},
\end{equation}
where $I_{i,j}$ is the original pixel intensity at location $(i,j)$,
$P_{i,j}$ is the intensity of the final panorama at location $(i,j)$, $\varepsilon$ is a weight,  $\nabla P_{i,j}$ is the gradient of the final panorama, $g_{i,j}$ is the gradient at location $(i,j)$ in the gradient map $g$(by putting the gradient of each stream $g^i$ together):
\begin{equation}\label{equ:gmap}
g = \sum\limits_{i = 1}^n {{g^i}M{'_i}}
\end{equation}
 Unlike in the original Poisson blending approach, $(i,j)$ now ranges over the whole image, so all the streams change the pixel value.
Tanaka et al~\cite{Tanaka:2012:SIC:2407156.2407173} solve this equation in the frequency domain :
\begin{equation}\label{equ:mpb2}
P{^T_{i,j}} = \frac{{v{^T_{i,j}} - \varepsilon u{^T_{i,j}}}}{{d{^T_{i,j}} - \varepsilon }},
\end{equation}
where $P{^T_{i,j}}$ is the DCT of each pixel in the final panorama, $v{^T_{i,j}}$ is the DCT of the Laplacian of the image (by putting the Laplacian of each stream together as in the Equation~\ref{equ:gmap}),
 $u{^T_{i,j}}$ is the DCT of the original intensity image,
  and $d{^T_{i,j}}$ is the DCT of the Laplacian operator. The final panorama is obtained by computing the inverse DCT of $P{^T_{i,j}}$.

\subsection{Intensity changes}
\begin{figure*}[t!]
\centering
\includegraphics[width=0.32\linewidth]{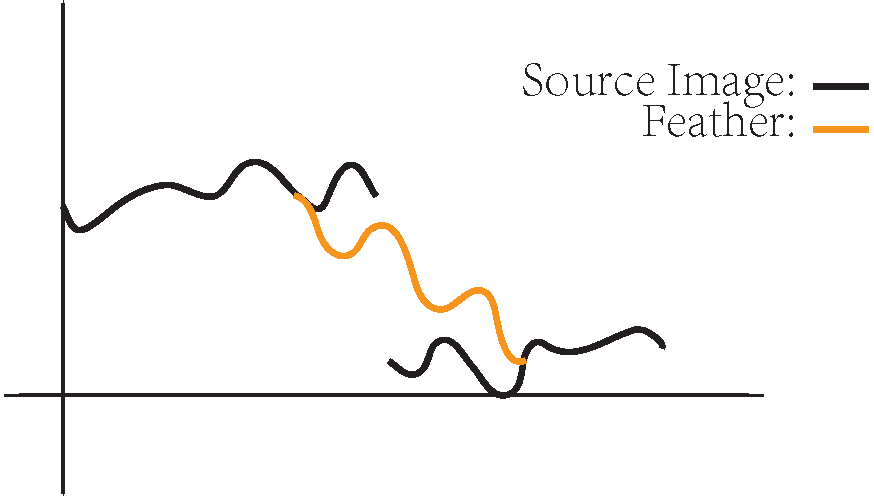}
\includegraphics[width=0.32\linewidth]{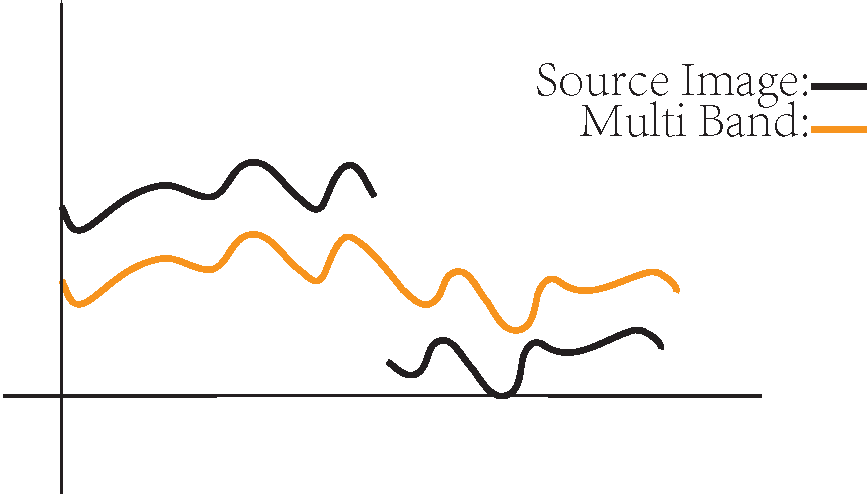}
\includegraphics[width=0.32\linewidth]{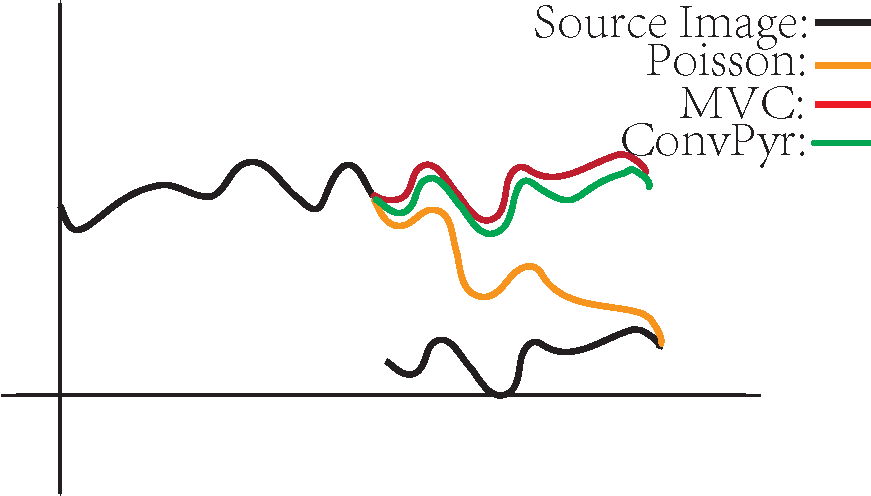}
\includegraphics[width=0.32\linewidth]{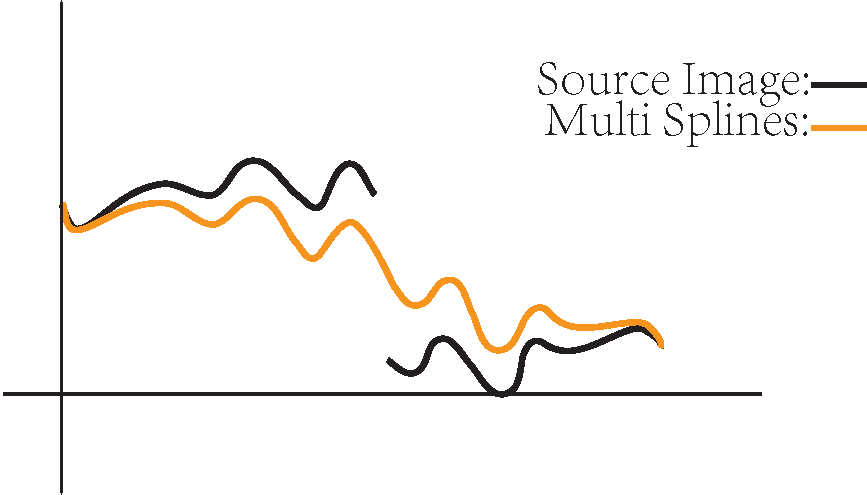}
\includegraphics[width=0.32\linewidth]{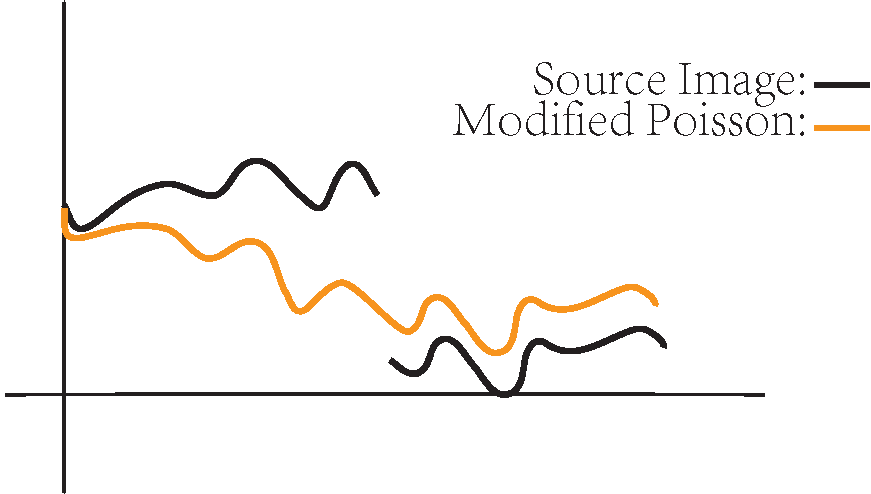}
\caption{One dimensional illustration of results produced by different blending algorithms.}
\label{fig:trend_comparison}
\end{figure*}

Since different blending algorithms have different formulations, they affect the pixel intensities in the result in different ways. We illustrate the trends
 of pixel intensity changes of different blending algorithms in Figure~\ref{fig:trend_comparison} and give a real world case in Figure~\ref{fig:result1}.
 Feather blending linearly blends the images in the overlapped regions, so other regions remain unchanged. Multi-band blending blends the images everywhere at different frequencies, causing intensities to be averaged across the whole image. Since MVC blending approximates Poisson blending, and convolution pyramid blending further approximates MVC blending, the regions to be blended changes in intensity to fit the anchor region in both of these two algorithms.
Multi-spline blending uses splines to approximate the offset map,so  lighting inconsistency  is obvious along the boundary seams especially if the input scenes are not well aligned  (see Figure~\ref{fig:result3}, 2nd row, 2nd column). Modified Poisson blending tries to preserves the original intensities
 as well as the gradient fields
  of the blended region, so it produces rather different results to all the other algorithms. Thus, MVC blending, modified poisson blending and convolution pyramid blending are sensitive for anchor stream choosing while feather blending, multi-band blending and multi-spline blending produce same blending results given arbitrary blending order.
\section{Experiments}
\label{sec:exp}
Our experiments were performed on a PC with an Intel Xeon E5-2620 2.0GHz CPU with 32GB memory, and an nVidia GTX 970 GPU with 4GB memory; the bandwidth between PC memory and  GPU memory was 4GB/s. The  blending algorithms were implemented in C++, while GPU implementations used CUDA.

\subsection{Efficiency}
We initially considered the theoretical time complexity of these 6 representative algorithms. Since it only computes a linear combination for each pixel, the complexity of feather blending is $O(n)$ where $n$ is the number of pixels. Multi-band blending also  complexity $O(n)$, as the extra levels only multiply the number of pixels to process by a constant factor. MVC blending requires target region triangulation and adaptive boundary sampling, with $O(m)$ cost for evaluating the membrane,  where $m$ is the number of pixels along the boundary; this is typically $O(\sqrt n)$. Since the last step interpolates the membrane values to all $n$ pixels, the total cost $O(n)$. Convolution pyramid blending uses small kernels to approximate a large kernel, so its complexity is again $O(n)$. Multi-spline blending needs to solve an $O(n/{s^2})$ linear system where $s$ is the sampling space of the spline, which thus has complexity higher than $O(n)$. Modified Poisson blending finds pixels in the frequency domain with complexity $O(n\log (n))$.

We experimentally measured the time required by each blending algorithm, as well as the memory it used.
The resolutions of the output blended videos were 4000$\times$2000. 
Note that the time and memory costs only depend on the resolution of the input videos and the shape of the mask, and not on the video content, so we just used one scene for this experiment. I/O times  as well as precomputation times were not considered, as we are interested in how suitable each method is for continuous realtime operation.
For each algorithm, Table~\ref{tab:t1} gives times(per frame)
and memory costs, both for CPU implementation, and where appropriate, GPU implementation.
The results show that when using a GPU with sufficient memory, multi-band blending, MVC blending, feather blending and convolution pyramid blending can  achieve realtime performance. 

\subsection{Visual quality}
Secondly, we both objectively and subjectively evaluated the blended videos produced by these algorithms.
We used 12 representative scenes from our benchmark for evaluation. We do not use all the scenes because we want to limit the subjective evaluation for each candidate within 20 minutes(we have 6 algorithms for comparison). Detail of the scenes that have been evaluated are presented in the supplementary material.

\subsubsection{Objective evaluation}
\begin{figure}[t!]
\begin{tabular}{c c c}
  \includegraphics[width=0.31\linewidth]{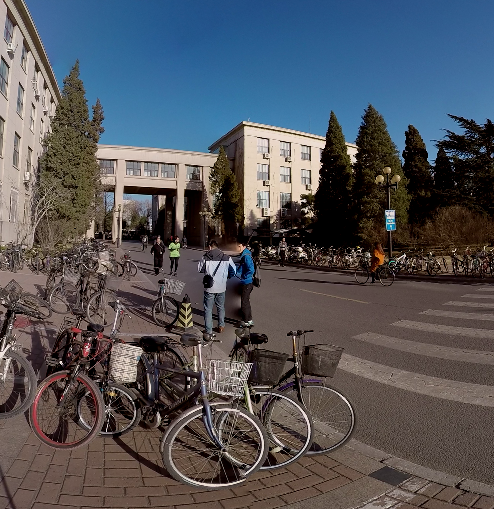} & \hspace{-0.15in} \includegraphics[width=0.31\linewidth]{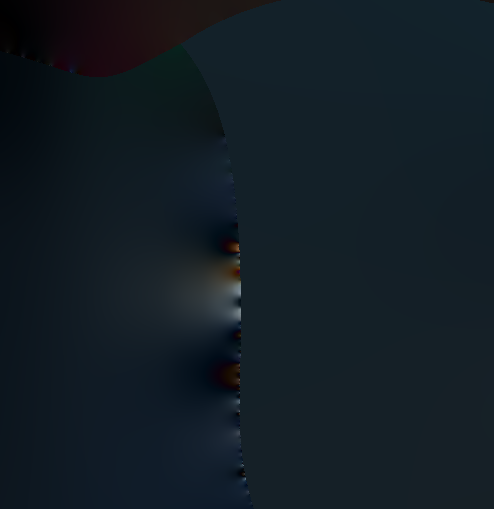} & \hspace{-0.15in} \includegraphics[width=0.31\linewidth]{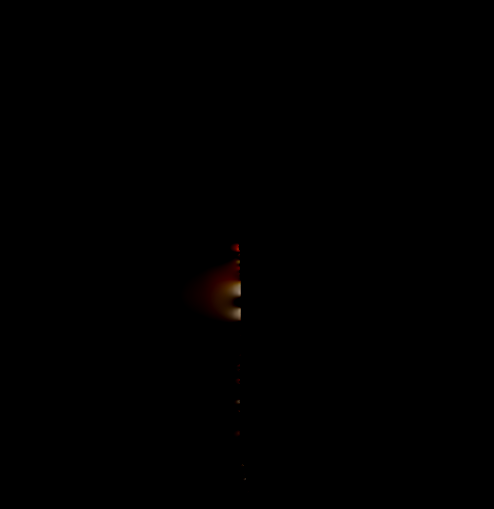}\\
  \includegraphics[width=0.31\linewidth]{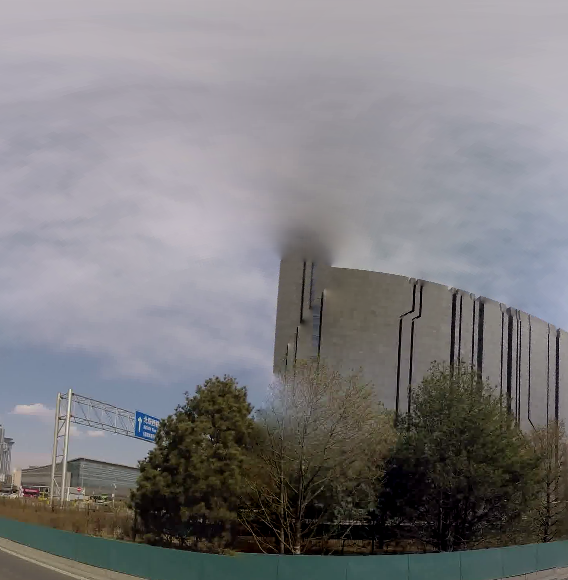} & \hspace{-0.15in} \includegraphics[width=0.31\linewidth]{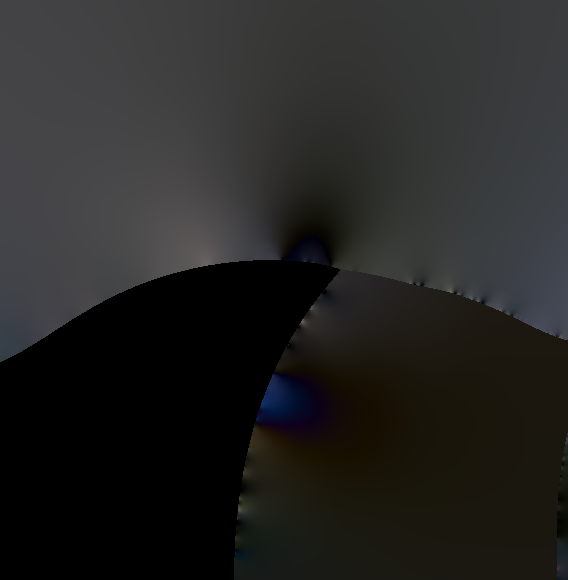} & \hspace{-0.15in} \includegraphics[width=0.31\linewidth]{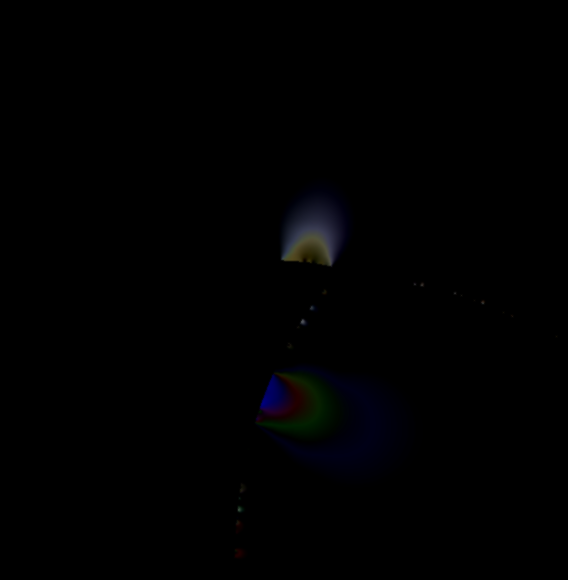}\\
  (a)  & (b) & (c)
\end{tabular}
\centering
   \caption{(a) blended result with MVCB(cropped from the panorama) (b) offset map (c) bleeding map}
\label{fig:bleeding}
\end{figure}

\begin{figure*}[t!]
\centering
\includegraphics[width=0.32\linewidth]{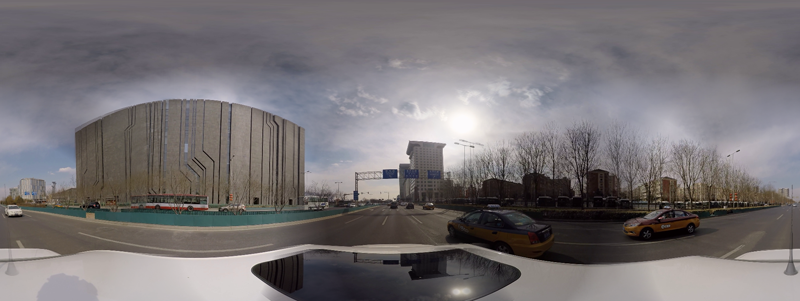}
\includegraphics[width=0.32\linewidth]{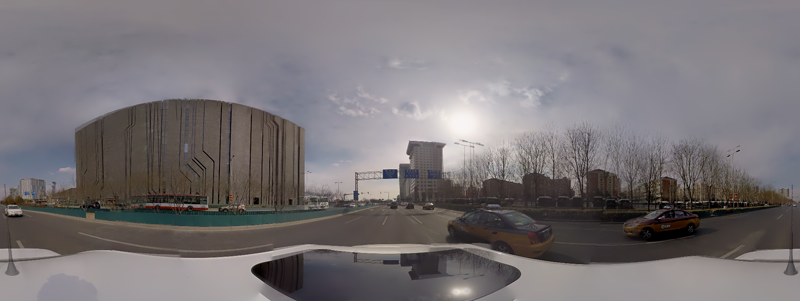}
\includegraphics[width=0.32\linewidth]{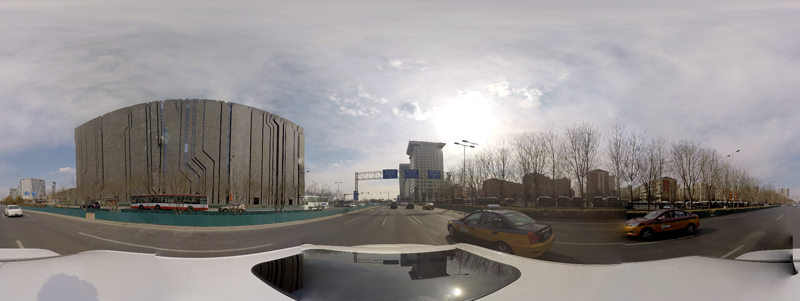}
\includegraphics[width=0.32\linewidth]{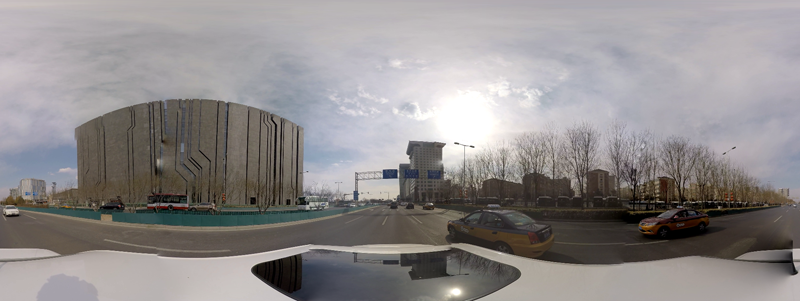}
\includegraphics[width=0.32\linewidth]{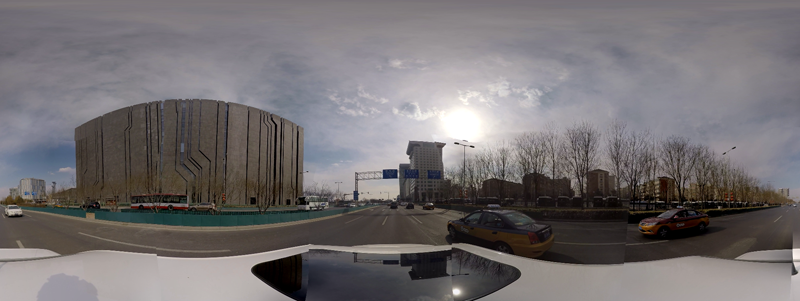}
\includegraphics[width=0.32\linewidth]{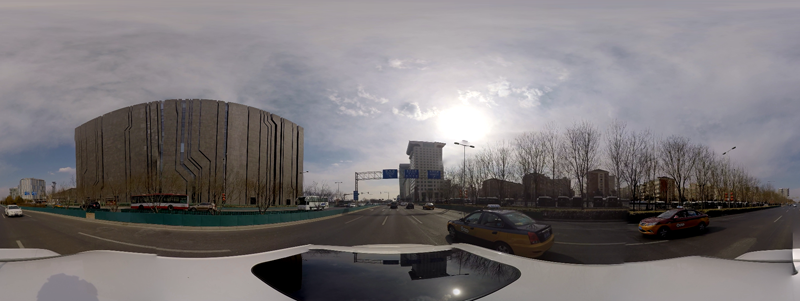}
\caption{An challenge scene with capturing device moving. Row 1, column 1-3 are results of feather blending, multi-band blending, MVC blending and row 2, column 1-3 are the results of convolution pyramid blending, multi-spline blending and modified poisson blending.}
\label{fig:result3}
\end{figure*}

Image and video quality assessment methods can be classified into double ended~\cite{4775883} and single-ended~\cite{Lin2011297}  approaches. Double-ended approaches such as PSNR (peak signal-to-noise rRatio) and SSIM (structural similarity) require an original image or video as a reference, but in video blending  there is no such ground truth. Thus, our objective metric for blended videos focuses on the most obvious blending artefacts. Specifically we design a metric to quantize the amount of bleeding.
Poisson blending and its variants suffer from bleeding(looks like a particular color leaking to its surroundings) artefacts.
 Examples of bleeding and the corresponding offset maps are illustrated in Figure~\ref{fig:bleeding}. The latter show that bleeding appears in regions with severe color differences to the color that appears most in other regions.
 To quantify the degree of bleeding, we first calculate an energy map by calculating the absolute values of the offset map:
We next calculate calculate the bleeding map based on the energy map:
 \begin{equation}\label{equ:relu}
B(p) = \max (0,p - \alpha \frac{{{E_h}}}{{{A_h} + \delta }}).
 \end{equation}
In the above equation $A_h$ is the number of non-zero values of the binarized energy map using Otsu's method~\cite{4310076}, and $E_h$ is the sum of the energy of the non-zero positions on the binarized map. $\alpha$ is a weight  set it to 2 in our experiment. This weight is used to truncate with a high peak value.
Given the bleeding map, we define the total amount of bleeding per frame as
\begin{equation}\label{equ:bleeding_degree}
{P_B} = \sum {{B(p)}^2},
\end{equation}
This quantity is then averaged over all frames. For the scene in Figure~\ref{fig:result3},
 the bleeding degrees for MSB, MPB, CPB and MVCB are 1.94, 23.34, 81.00 and 142.90 respectively. This implies that for not well aligned videos CPB and MVCB have more severe degree of bleeding compared with MSB and CPB. This is because CPB and MVCB have more strict boundary conditions than MSB and MPB. Further results are provided in the supplementary material.


\subsubsection{Subjective evaluation}
Following~\cite{itubt500}, we conducted our experiments in a standard office environment and used integers from 1 to 5 inclusive for subjective scoring, higher meaning better visual quality. For the 12 test videos we picked 10~s from each scene
and provided the results of 7 different algorithms (the above 6, plus simple stitching without blending), giving each participant 840~s of video to view ($12 \times 10 \times 7$).  The 20 participants included  7  majoring in computer vision and computer graphics,  7  students from other research areas such as computer networks and data mining, and 6 students studying other subjects.

We first gave each participant a familiarisation session before evaluation, showing several typical scenes and their blending results, to help participants become more aware of issues in visual blending quality. These were annoted with an expert's remarks on the video such as `there is an obvious seam and the color is not very consistent near the seam' or `there is a flicker in the moving object'. Several kinds of artefact were also listed.  After the training session, the scenes showing results of different blending algorithms were presented to each participant in a random order.

After the experiment, results were filtered~\cite{HantaoLiu2010} to  reject outlier evaluations and individuals. Results more than two standard deviations from the mean score for that test were considered to be outliers; an individual was an outlier if 1 out of 3 of his scorings were outliers. This caused 1 participant to be rejected.

After data filtering, for each algorithm and each scene, we calculated the mean scores for the remaining 19 participants.
 We also calculated the mean and variance for each algorithm over all 12 scenes, which reflect the average performance and stability of each algorithm respectively. The results are presented in Table~\ref{tab:t4}.

\begin{figure*}[t!]
\centering
\includegraphics[width=0.32\linewidth]{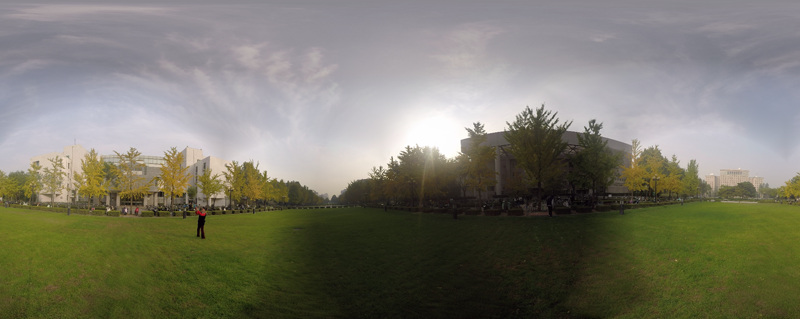}
\includegraphics[width=0.32\linewidth]{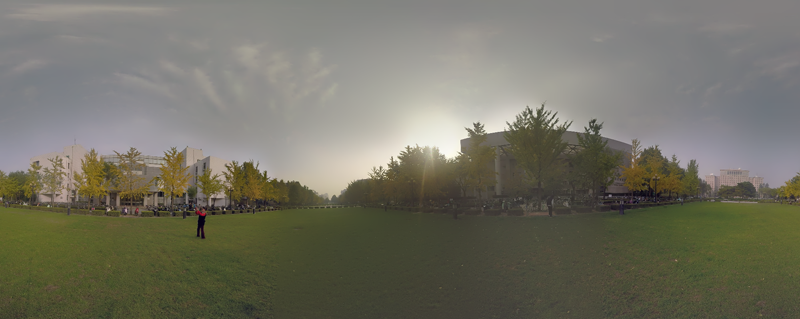}
\includegraphics[width=0.32\linewidth]{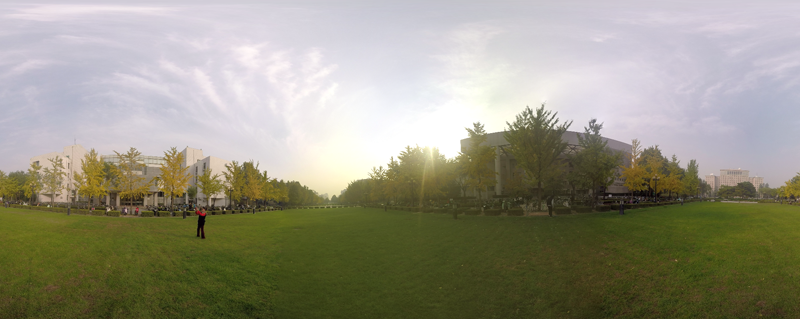}
\includegraphics[width=0.32\linewidth]{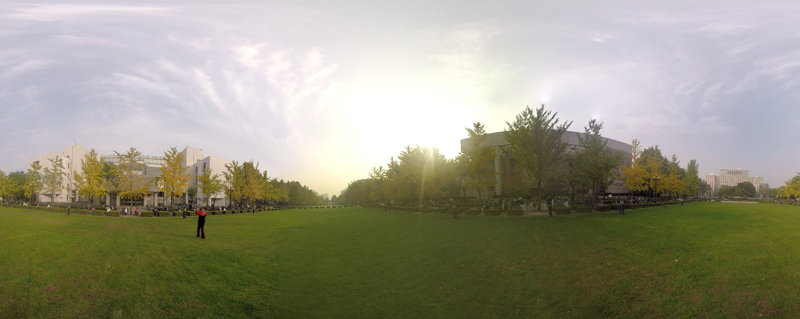}
\includegraphics[width=0.32\linewidth]{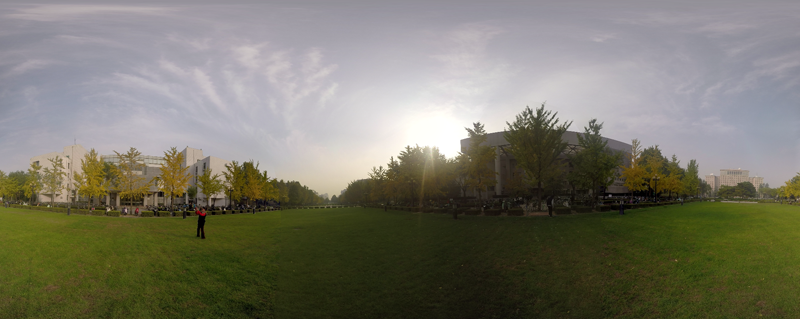}
\includegraphics[width=0.32\linewidth]{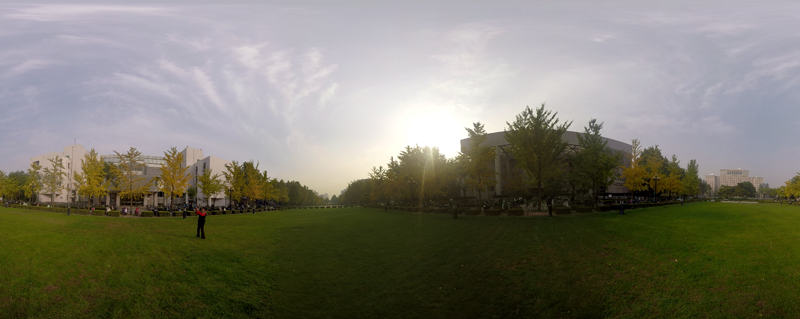}
\caption{An outdoor scene with obvious illuminance differences for different streams. Row 1, column 1-3 are results of feather blending, multi-band blending, MVC blending and row 2, column 1-3 are the results of convolution pyramid blending, multi-spline blending and modified poisson blending.}
\label{fig:result1}
\end{figure*}

 The main problem with feather blending is that it only blends the content in the overlapping region, and when  obvious illuminance differences exist, as in Figure~\ref{fig:result1}, it produces poor results. The result of multi-band blending is slightly blurred, and ghosting artefacts exist in scenes that are not well aligned.
  Multi-spline blending, MVC blending and convolution blending produce similar results, and they are not very stable because they are sensitive to misalignment. Modified Poisson blending generates higher quality results and is more stable.


\begin{table*}
\caption{Mean and variance of subjective score (higher is better) over 12 scenes, for each algorithm.}
\centering
\label{tab:t4}
\begin{tabular}{|c|c|c|c|c|c|c|c|}
  \hline
    & FB  & MBB  & MSB & MVCB & CPB & MPB & No Blending \\
    \hline
  Mean & 2.95 & 3.11 & 3.05 & 3.00 & 2.95 & 3.53 & 1.47  \\
  \hline
  Variance & 1.18  &  0.97 & 0.81  &  0.97 & 0.83 &0.66 & 0.57 \\
  \hline
\end{tabular}
\end{table*}

\begin{figure*}[t!]
\centering
\includegraphics[width=0.32\linewidth]{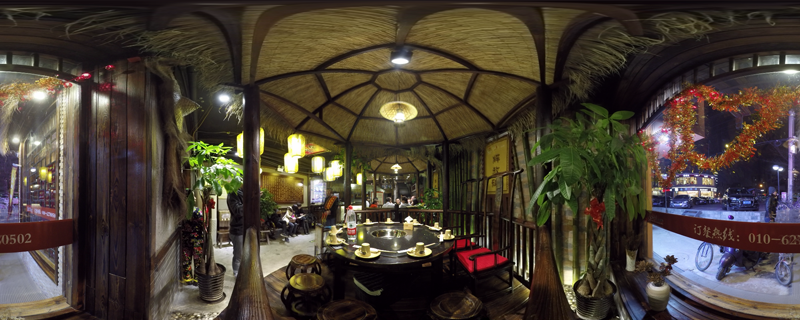}
\includegraphics[width=0.32\linewidth]{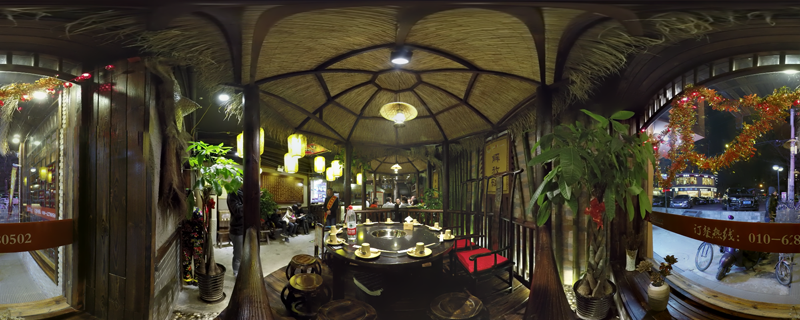}
\includegraphics[width=0.32\linewidth]{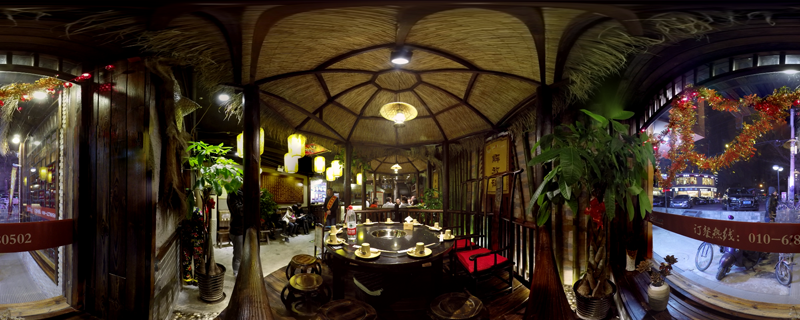}
\includegraphics[width=0.32\linewidth]{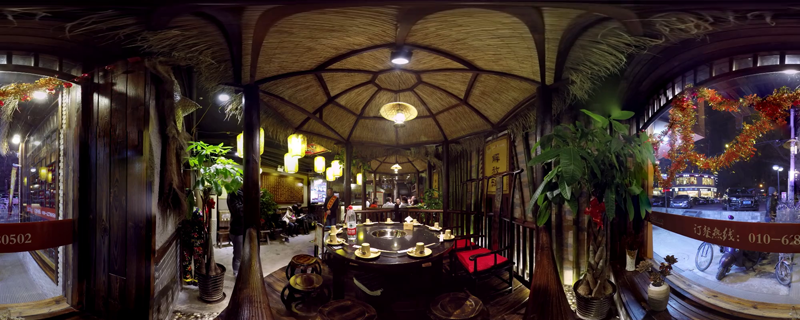}
\includegraphics[width=0.32\linewidth]{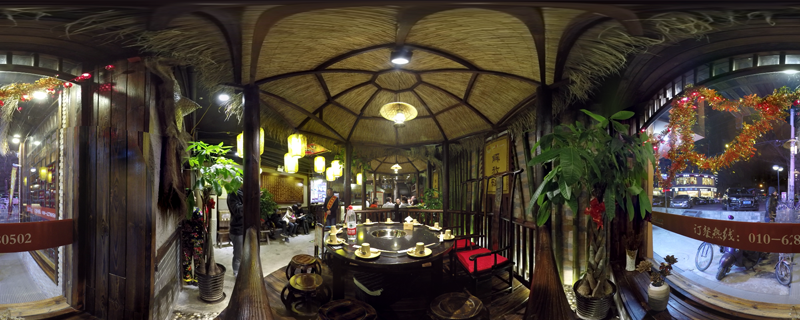}
\includegraphics[width=0.32\linewidth]{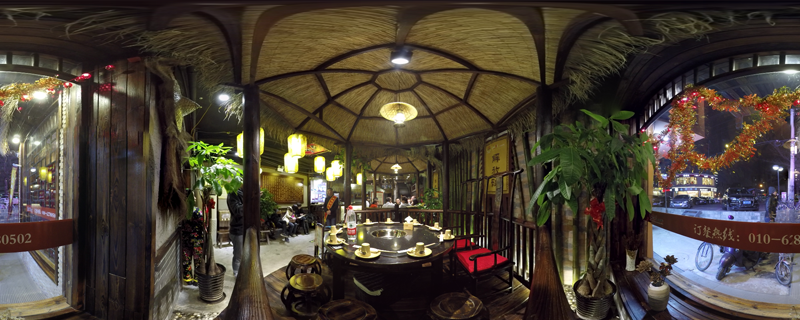}
\caption{An indoor scene. Row 1, column 1-3 are results of feather blending, multi-band blending, MVC blending and row 2, column 1-3 are the results of convolution pyramid blending, multi-spline blending and modified poisson blending.}
\label{fig:result2}
\end{figure*}

\section{Conclusions}
\label{sec:conclusion}
We have compared the performance and visual quality of 6 blending algorithms when used for realtime 4K video blending for a variety of scenes. Simple approaches such as FB and MBB are very efficient in GPU, but they can not produce very high quality blending results. The main problem of MVCB and CPB is that they are too sensitive to boundary condition, and suffer from bleeding even for just 1 pixel's misalignment. MSB suffers less from bleeding compared with MVCB and CPB, but obvious lighting inconsistency exists when it is not well aligned.

Our experiments show that modified Poisson blending performs surprising well on various scenes. However, it is not as efficient as some other approaches. More work is needed to improve the efficiency of modified Poisson blending by use of approximation techniques.
{\small
\bibliographystyle{ieee}
\bibliography{egbib}
}

\end{document}